%% file: main.tex
\documentclass[10pt,twocolumn,letterpaper]{article}

\usepackage{iccv}              % To produce the CAMERA-READY version
\usepackage{float}
\usepackage{algorithm}
\usepackage{algorithmic}
\usepackage{caption}
\usepackage{dsfont}
\usepackage{tabularx}
\usepackage{diagbox}
\usepackage[accsupp]{axessibility}
\usepackage{multirow}
\usepackage{makecell}
\usepackage{bbding} %对号错号
\input{preamble}

\usepackage{threeparttable}
\usepackage{enumitem}
\usepackage{graphicx}
\definecolor{cvprblue}{rgb}{0.21,0.49,0.74}
\usepackage[pagebackref,breaklinks,colorlinks,citecolor=cvprblue]{hyperref}

\def\paperID{11924} % *** Enter the Paper ID here
\def\confName{ICCV}
\def\confYear{2025}
\title{PDE: Gene Effect Inspired Parameter Dynamic Evolution \\ for Low-light Image Enhancement}
\def\authorBlock{
	Tong Li$^{1} $ \qquad Lizhi Wang$^{2}$ \qquad Hansen Feng$^{1}$ \qquad Lin Zhu$^{1}$ \qquad Hua Huang$^{2}$ \\
	$^{1}$ School of Computer Science and Technology, Beijing Institute of Technology \\
	$^{2}$ School of Artificial Intelligence, Beijing Normal University \\
}

\author{\authorBlock}
\begin{document}
\maketitle

%%%%%%%%% ABSTRACT
\begin{abstract}
Low-light image enhancement (LLIE) is a fundamental task in computational photography, aiming to improve illumination, reduce noise, and enhance image quality.
While recent advancements focus on designing increasingly complex neural network models, we observe a peculiar phenomenon: resetting certain parameters to random values unexpectedly improves enhancement performance for some images. 
Drawing inspiration from biological genes, we term this phenomenon the gene effect. The gene effect limits enhancement performance, as even random parameters can sometimes outperform learned ones, preventing models from fully utilizing their capacity.
In this paper, we investigate the reason and propose a solution.
Based on our observations, we attribute the gene effect to static parameters, analogous to how fixed genetic configurations become maladaptive when environments change.
Inspired by biological evolution, where adaptation to new environments relies on gene mutation and recombination, we propose parameter dynamic evolution (PDE) to adapt to different images and mitigate the gene effect.
PDE employs a parameter orthogonal generation technique and the corresponding generated parameters to simulate gene recombination and gene mutation, separately.
Experiments validate the effectiveness of our techniques.
The code will be released to the public.
\end{abstract}

%%%%%%%%% BODY TEXT

\input{1_introduction}

\input{2_Related}

\input{2_motivation_and_rethinking}

\input{3_method}

\input{4_experiment}

\input{6_conclusion}

{
	\small
	\bibliographystyle{ieeenat_fullname}
	\bibliography{references}
}

\end{document}

%% file: preamble.tex
\usepackage[dvipsnames]{xcolor}

%% file: 1_introduction.tex
\section{Introduction}

%%%%%%%%%%%%%%%%%%%%%%%%%%%%%%%%%%%%%%%%%%%%%%%%%%%%%%%%%%%%%%%%%%%%%%%%第3版
Low-light image enhancement (LLIE) aims to improve illumination, reduce noise, and enhance image quality of the low-light images~\cite{RetinexNet}, which is a fundamental task in computational photography~\cite{LLIE_suervey} and an essential step for high-level computer vision tasks~\cite{Detection,Detection3}. The diverse degradation impose significant challenges for LLIE methods~\cite{zhang2018degradations,xu2020degradations}. 
Achieving high-quality results with suitable color and brightness has been a longstanding objective in this field~\cite{xia2023diffir,DiffLL}.

The mainstream approaches train neural network models to map low-light images to high-light images~\cite{Retinexformer,LLIE_suervey}. 
In recent years, most research efforts focused on customizing complex neural network models, evolving from CNN-based models~\cite{xu2020degradations,Enlightengan}, Transformer-based models~\cite{Retinexformer,Restormer}, to Mamba-based models~\cite{retinexmamba,mambair}.

\input{Figure/teaser}

However, we have observed a counterintuitive phenomenon in existing models, which we refer to as the “gene effect”, as illustrated in Figure~\ref{fig:teaser}.
Surprisingly, resetting certain parameters to random values—which would typically degrade enhancement performance—can instead improve enhancement for some images.
We term this peculiar phenomenon gene effect, drawing inspiration from biological genes, where random mutations benefit some individuals while harming others.
The gene effect is harmful to some images, as even random parameters can achieve better enhancement performance, hindering the model from fully utilizing the enhancement capacity and creating a significant obstacle to further performance improvements.
Similar to biological genes that cannot simply be deleted, gene effect-related parameters cannot be directly pruned. 
Thus, mitigating the gene effect becomes a particularly intriguing challenge.

\input{Figure/bio}

%%%%%%%%%%%%%%%%%%%%%%%%%%%%%%%%%%%%%%%%%%%%%%%%%%%%%%%%%%%%%%%%%%%%%%
As far as we know, no similar phenomenon has been reported in LLIE or other related fields. 
In this paper, we investigate the reason and propose a solution, to illustrate and mitigate the observed gene effect phenomenon, through observations and biological gene mechanisms.

The investigation into the reason for the gene effect begins with the observation, that dynamic parameters exhibit weaker gene effects than static parameters. 
To better understand this observation, we draw insights from biological gene evolution.
In biological gene evolution, static genetic configurations are well-suited for stable environments but become maladaptive and even harmful when ecological conditions shift. 
Similarly, we attribute the reason for the gene effect to the static parameters. 
Currently, LLIE models apply static parameters learned after training to all input images, which are suitable for some images but become maladaptive and even more harmful than random parameters when facing particular images.

The investigation into the solution to the gene effect begins with insights from biological gene evolution. 
To adapt to new environments, biological individuals rely on gene mutation and gene recombination to dynamically evolve genes~\cite{Dobzhansky}, as shown in Figure~\ref{fig:bio}.
Here, we propose a parameter dynamic evolution (PDE) method to adapt different images to mitigate gene effect.

Inspired by gene mutation, which generates new genes and traits to adapt to new environments, 
PDE employs dynamic parameters to simulate parameter mutation and mitigate the gene effect.
However, the current dynamic parameter mechanism also exhibits the gene effect. The dynamic parameter mechanism learns multiple candidate parameters but sometimes relies on a single candidate parameter and degrades to static parameters~\cite{hssayni2022redundancy,Robustness}.

Inspired by gene recombination, where orthogonal genetic information prevents the excessive expression of similar genes~\cite{Morgan}, PDE employs the parameter orthogonal generation (POG) technique to generate dynamic parameters and avoid degradation to static parameters. 
Specifically, POG learns orthogonal basis embeddings of parameters and dynamically generates suitable parameters based on the orthogonal bases.

Our contributions are summarized as follows:
\begin{itemize}[leftmargin=0.8cm]
	
	\item To the best of our knowledge, we are the first to identify and illustrate the gene effect in LLIE.
	\item We propose the PDE method to mitigate the gene effect, primarily relying on the POG technique.
	\item Experiments show our method mitigate the gene effect while improving the performance of LLIE.
\end{itemize}

%% file: Figure/teaser.tex
\begin{figure}
	\centering
%	\vspace{-2em}
	\includegraphics[width=\linewidth]{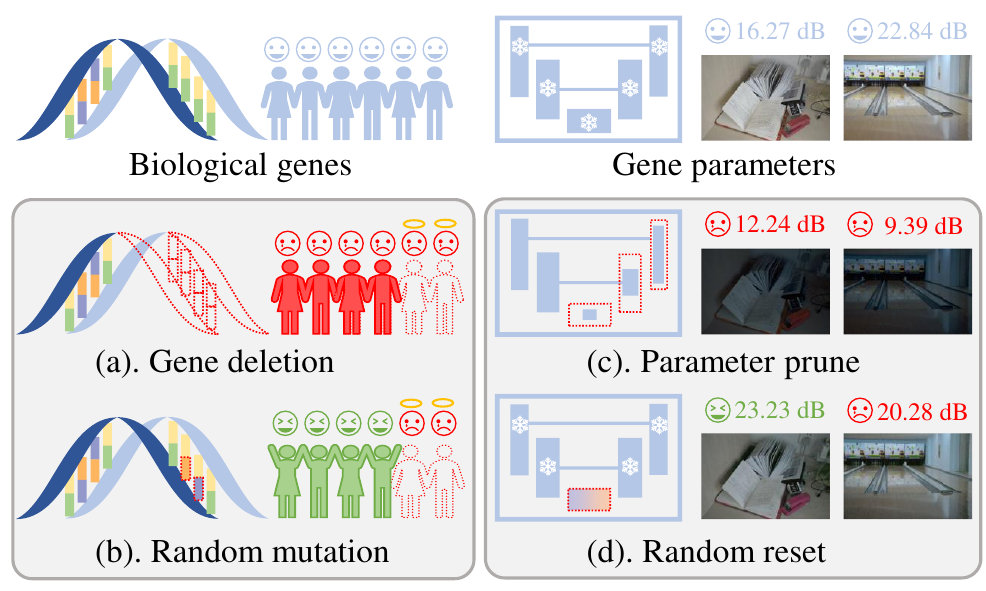} 
	\caption{\textbf{Overview of gene effect.} Resetting certain parameters to random values can even improve enhancement performance for some images.
		Inspired by biological genes, where random mutations benefit some individuals while harming others, we name this peculiar phenomenon “gene effect”.
	}
%	\vspace{-1em}
	\label{fig:teaser}
\end{figure}

%% file: Figure/bio.tex
\begin{figure}
	\centering
	\includegraphics[width=\linewidth]{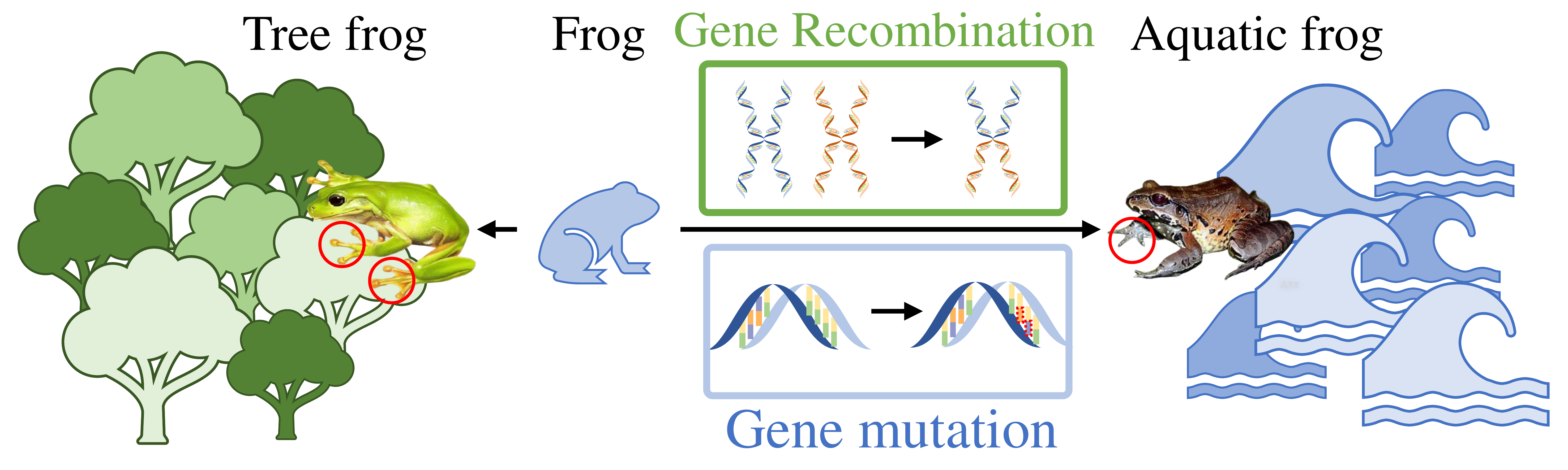} 
	%	\vspace{-1em}
	%	\caption{\textbf{Dynamic convolution.} 
		%		}
	\caption{\textbf{Biological gene evolution.} 
		Biological individuals rely on gene mutation and gene recombination to dynamically evolve gene and adapt to new environments.
		For gene mutation, new genetic information are generated. 
		For gene recombination, different genetic information are changed. 
%		Inspired by gene mutation, we propose parameter dynamic evolution (PDE) to mitigate gene effect. Inspired by gene recombination, we further polish PDE method.
Inspired by biological gene evolution, we propose PDE, which employs a parameter orthogonal generation technique and the corresponding generated parameters to simulate gene recombination and gene mutation, separately.
	}	
	\label{fig:bio}
%	\vspace{-2em}
\end{figure}

%% file: 2_Related.tex
\section{Related work}

\subsection{Low-light image enhancement}
Traditional low-light image enhancement methods focus on employing image priors, for example,
histogram equalization~\cite{HE4}, gama correction~\cite{GC3} and Retinex theory~\cite{RT2,RT3}.
Histogram-based methods~\cite{HE3,HE4,HE5,HE6} and gamma-based methods~\cite{GC1,GC3} focus on directly enhancing illumination. These methods typically rely on empirically derived prior knowledge to achieve brightness adjustments. Retinex-based methods~\cite{RT2,RT3,RT4,NPE,LIME} are grounded in human cognition theories, dividing the image into an illumination map and a reflectance map. These  Retinex-based methods generally require enhancing the illumination map while simultaneously denoising the reflectance map.
However, the ability of these traditional methods in complex degradation conditions is limited. 

With the
development of deep learning, learning-based methods become the mainstream methods. Current mainstream approaches train neural networks to map low-light images to high-light images~\cite{Detection,Detection3}. In recent years, most research efforts focused on refining the neural network architectures~\cite{retinexmamba,Retinexformer,Restormer,MIRNet,IACG}. 
The low-light image enhancement methods have evolved from CNN-based methods~\cite{RetinexNet,Kind,Kind_plus,ZeroDCE,CNN3,DeepUPE,DeepLPF,Enlightengan,CNN4} to Transformer-based methods~\cite{LLformer,Restormer,SNRNet,MIRNet}, Diffusion-based methods~\cite{DiffLL,GSAD,Diffretinex,GLARE} and Mamba-based methods~\cite{retinexmamba,mambair,LLEMamba,MambaLLIE}. As networks become more advanced, the enhancement performance improves. However, the significant model redundancy within these methods prevents further performance improvement.

\subsection{Dynamic parameter}
The existing method for dynamic parameter generation is dynamic convolution~\cite{dynamicconvolution,condconvolution}, which learns multiple candidate convolutional parameters and dynamically weights the candidate parameters based on the input image. Without specific constraints, the dynamic convolution easily learns similar or relevant candidate convolutions~\cite{hssayni2022redundancy}, as even initialization~\cite{Robustness} can lead to various correlations. 
As far as we know, there are no LLIE methods based on dynamic convolution.
The few other image restoration methods~\cite{ADFNet,dynamicsr,dynamicie,dynamicsr2} that use dynamic convolution have shown limited performance.

\input{Table/tab_1_obervation_3}
\input{./Figure/1_intro_problem}
\subsection{Biological gene evolution}
The study of biological gene evolution has profoundly shaped the understanding of adaptability, diversity, and survival mechanisms in nature. 

Darwinian Natural Selection~\cite{Darwin} establishes that beneficial traits are naturally selected, while the Modern Synthesis~\cite{Dobzhansky} integrates genetics with natural selection, emphasizing gene mutation and gene recombination as key drivers of gene evolution. 
The Neutral Theory~\cite{Kimura} further proposes that most gene mutations are neutral, with genetic drift playing a significant role in evolution.
In summary, most studies agree that biological gene evolution helps species respond to environmental changes.
%Evolutionary dynamics~\cite{Wagner} shape how species respond to environmental changes.

Biological gene evolution arises from several mechanisms. Gene mutation introduces random changes in DNA sequences \cite{Muller}, while chromosomal recombination ensures genetic diversity by mixing parental genes during meiosis \cite{Morgan}. Beyond these classical processes, there are also new theories. For example, horizontal gene transfer \cite{Ochman} allows certain organisms to acquire foreign genes, enabling rapid adaptation.
These studies have established biological gene evolution as a fundamental process driving biodiversity, providing insights into how genetic structures adapt to dynamic environments.
%﻿

%% file: Table/tab_1_obervation_3.tex
\begin{table}
	\begin{center}
		\setlength{\tabcolsep}{4.2pt}
		\caption{POI represents the percentage of images that get better results when resetting the well-trained parameters with the random values. Over 30\% images achieve better enhancement performance.
		}		  
		\vspace{-1em}
		\small
		\addtolength{\tabcolsep}{4pt}
		\begin{tabular}{l c c c c c c}
			\toprule
			NOL &1st &2nd &3rd &4th &5th &6th   \\
			\midrule
			POI & 40$\%$& 33$\%$&33$\%$&27$\%$& 27$\%$& 33$\%$  
			\\
			\bottomrule
		\end{tabular}
		\label{table:observation3}
		\vspace{-1em}
	\end{center}
\end{table}

%% file: Figure/1_intro_problem.tex
\begin{figure}[t]
	%	\vspace{-1em}
	\centering
	\renewcommand{\arraystretch}{1} % 调整行高
	\setlength{\tabcolsep}{1.5pt}
	\scalebox{0.97}{
		\small
		\begin{tabular}{c c c c}
			Input &
			Well-trained &
			Random &
			Reference
			\\	
			\includegraphics[width=.115\textwidth]{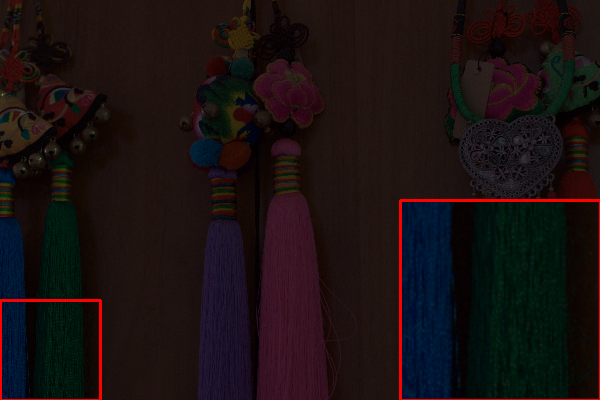} &   
			\includegraphics[width=.115\textwidth]{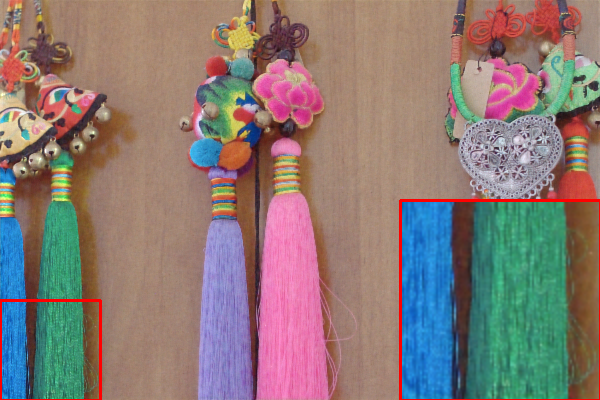} &
			\includegraphics[width=.115\textwidth]{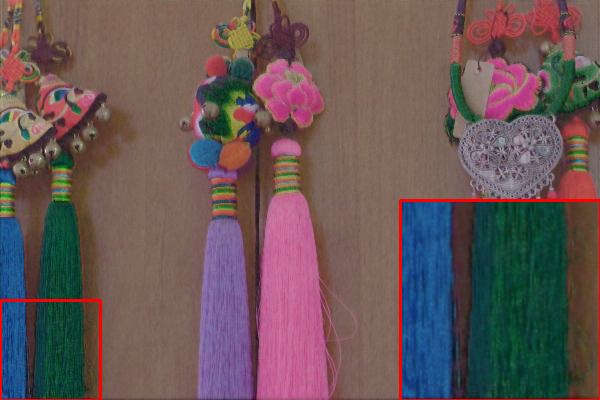} &
			\includegraphics[width=.115\textwidth]{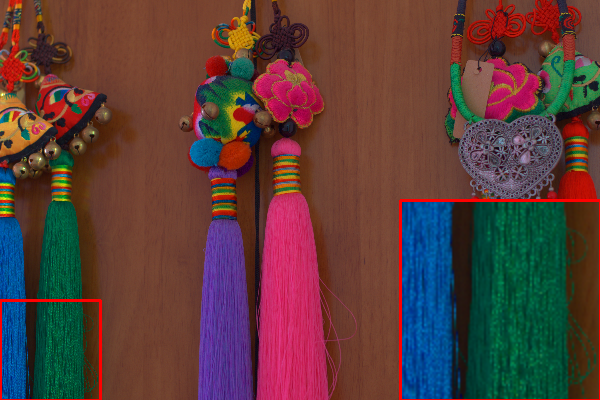}
			
			\\
			12.38 dB &
			12.72 dB &
			17.54 dB &
			PSNR
			\\
			
			\includegraphics[width=.115\textwidth]{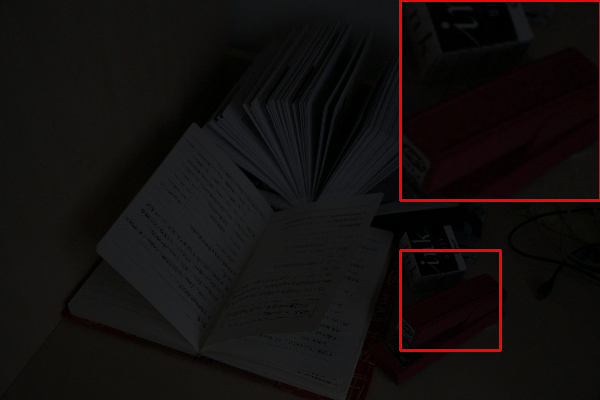} &   
			\includegraphics[width=.115\textwidth]{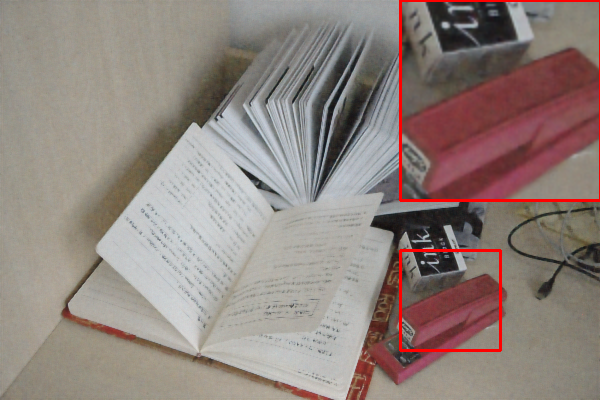} &
			\includegraphics[width=.115\textwidth]{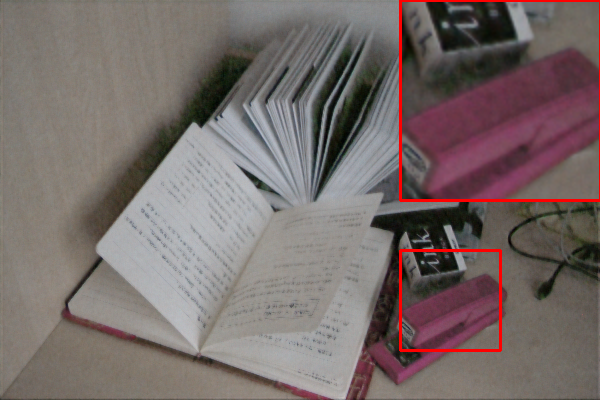} &
			\includegraphics[width=.115\textwidth]{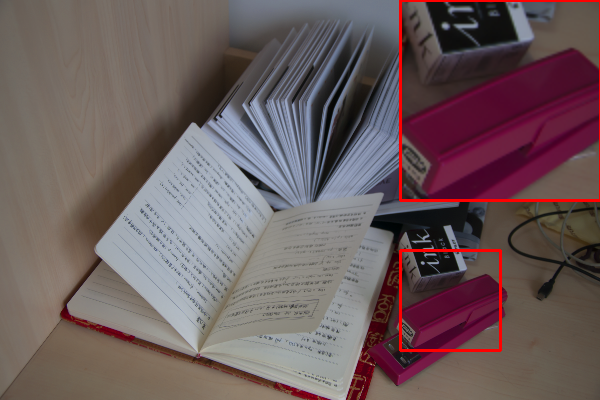}
			\\
			
			8.98 dB &
			16.27 dB &
			23.23 dB &
%			25.56 dB &
			PSNR
			\\

		\end{tabular}
	}
	%		\vspace{-1em}
	\caption{
		\textbf{Gene effect in low-light image enhancement models}.
	From left to right, the images are as follows: the low-light image, the image enhanced by the original well-trained Restormer~\cite{Restormer}, the image enhanced by the Restormer in which certain attention mechanism parameters have been reset to random values, and the reference image. 
	The image enhanced with the well-trained parameters exhibits overexposure and fading color, with only 12.72 dB.
	In contrast, the images enhanced with random parameters show even higher PSNR values, along with more accurate color and brightness. 
%	This peculiar phenomenon is named as gene parameters.
	}
	\label{fig:intro_problem}
%	\vspace{-1em}
\end{figure}

%% file: 2_motivation_and_rethinking.tex
\section{Observation and Discussion}

As far as we know, no similar phenomenon has been reported in LLIE or other related fields.
In this section, we analyze experimental observations and biological gene mechanisms to explore the reason for the gene effect and identify possible solutions to mitigate the gene effect.

The gene effect refers to the peculiar phenomenon where resetting certain parameters to random values unexpectedly improves enhancement performance for some images.

To demonstrate the gene effect phenomenon, we reset the well-trained parameters of Restormer~\cite{Restormer} to random values and evaluate the LLIE performance. (Additional experiments on other architectures are presented in Section~\ref{sec:Experiments}.) Surprisingly, we find that more than 30\% of images achieve better enhancement performance when the well-trained parameters are replaced with random values, as shown in Table~\ref{table:observation3} and Figure~\ref{fig:intro_problem}.

\input{Figure/dynamic_convolution}

\subsection{Reason} 
\label{sec:motivation_experiments}

\input{Table/tab_1_obervation_2}
\input{Figure/mapping}

The investigation into the reason for the gene effect begins with the observation, that dynamic parameters exhibit weaker gene effects compared to static parameters. 
Static parameters refer to the standard convolution layers, while dynamic parameters are generated by dynamic convolution~\cite{dynamicconvolution,condconvolution}. The dynamic convolution learns multiple candidate parameters and weights candidate parameters based on the input image characteristics, as shown in Figure~\ref{fig:dynamic}.
Resetting dynamic parameters caused a more serious performance drop than resetting static parameters, as shown in Table~\ref{table:observation2}, indicating that dynamic parameters exhibit weaker gene effects compared to static parameters.

To better understand this observation, we draw insights from biological gene evolution.
In biological gene evolution, static genetic configurations are well-suited for stable environments but become maladaptive or even harmful when ecological conditions shift~\cite{Kimura,Dobzhansky,Darwin}.
%Similarly, in LLIE models, static parameters may perform well for certain images but become maladaptive or even harmful when applied to others. 
Extending this analogy to LLIE models, we argue that static parameters, similar to static genes, perform well for certain images, similar to stable environments, but become maladaptive or even detrimental when applied to others.
A deeper analysis is as follows.
Specifically, LLIE training images do not follow the same distribution, as visually similar low-light images may be mapped to different high-light images (as shown in Figure~\ref{fig:mapping}), yet LLIE methods typically are composed of static parameters.
This forces the model to learn a general mapping — a compromised solution that adjusts all inputs toward similar brightness and color tones, limiting its adaptability to different images.
For example, as shown in Figure~\ref{fig:teaser}, the original model, pruned model, and reset model display different brightness preferences, separately. 
As a result, the general mapping may work well for some images, but even can not perform as well as random parameters for other images. 
In summary, we attribute the reason for the gene effect to the static parameters.

\input{./Figure/PDE}
\input{./Figure/POG}
\subsection{Solution} 
\label{sec:solution}

The investigation into the solution to the gene effect begins with insights from biological gene evolution. 
To adapt to new environments, biological individuals rely on gene mutation and gene recombination to dynamically evolve genes~\cite {Dobzhansky}, as shown in Figure~\ref{fig:bio}.

Inspired by the gene mutation of biological gene evolution, it is natural to employ dynamic parameters to perform parameter mutation to simulate gene mutation and deal with the gene effect. Here, we name it parameter dynamic evolution (PDE).
However, the current dynamic parameter mechanism also exhibits the gene effect, as the dynamic parameter mechanism sometimes relies on a single candidate parameter and degrades to static parameters~\cite{hssayni2022redundancy,Robustness}, as shown in Figure~\ref{fig:dynamic}.

Inspired by the gene recombination of biological gene evolution, we solve the previous problem.
%Inspired by the gene recombination of biological gene evolution, we further polish the PDE method.
For gene recombination, chromosome crossover recombination ensures the recombination of orthogonal genetic information~\cite{Morgan}, preventing the excessive expression of similar genes.
Thus for PDE, we propose a parameter orthogonal generation (POG) technique to generate dynamic parameters, which is a kind of dynamic parameter method based on orthogonal parameter information.

In summary, inspired by the gene mutation of biological gene evolution, we propose PDE to mitigate the gene effect.
PDE employs a parameter orthogonal generation technique and the corresponding generated parameters to simulate gene recombination and gene mutation, separately.

%% file: Figure/dynamic_convolution.tex
\begin{figure}
	\centering
	\includegraphics[width=\linewidth]{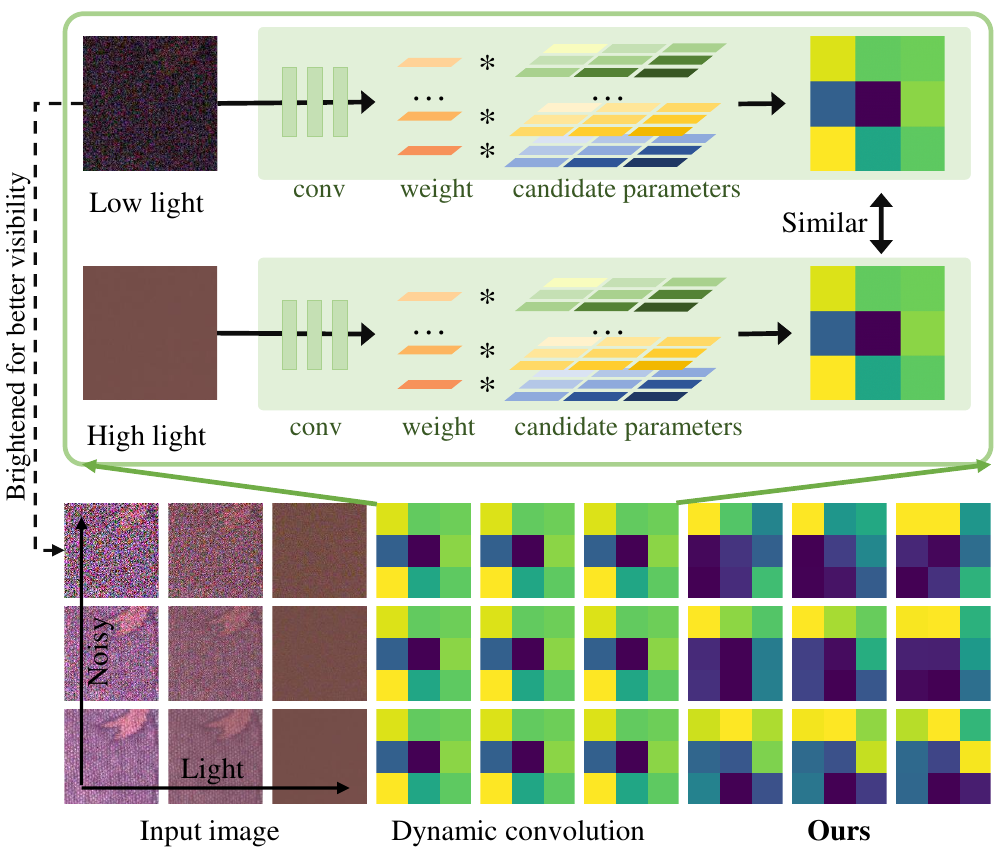} 
	%	\vspace{-1em}
%	\caption{\textbf{Dynamic convolution.} 
%		}
	\caption{\textbf{Comparison of the generated dynamic parameters.} 
		\textbf{The top row} presents the basic framework of current parameter mechanism, where the dynamic convolution employ convolutions to extract weights based on the input image to weight the candidate parameters.
		\textbf{The bottom} row presents comparison of the generated dynamic parameters. The dynamic parameters generated by ours for each row and column image exhibit gradual evolution processes, indicating the ability to recognize differences and understand similarities between these images.
		(The input low-light images in the comparison have been brightened for better visibility.)
	}	
	\label{fig:dynamic}
	\vspace{-0.5em}
\end{figure}

%% file: Table/tab_1_obervation_2.tex
\begin{table}
	\begin{center}
		\setlength{\tabcolsep}{1pt}
		\caption{The observation is that dynamic parameters exhibit weaker gene effects compared to static parameters.
		NOL represents the number of layer parameters to be reset to random values. 
		}	
		\vspace{-1em}	  
		\small
		\addtolength{\tabcolsep}{4pt}
		\begin{tabular}{l c c c c c c}
			\toprule
			NOL &1st &2nd &3rd &4th &5th &6th   \\
			\midrule
			Static& - 1.72 & + 0.05 &+ 0.30 & - 0.99 & - 0.01 & + 0.01 
			\\
			
			\midrule
			Dynamic &- 3.39&- 1.55&- 0.44&- 2.65&- 0.16&- 0.01\\

			\bottomrule
		\end{tabular}
		\label{table:observation2}
%		\vspace{-2em}
\vspace{-1em}
	\end{center}
\end{table}

%% file: Figure/mapping.tex
\begin{figure}
	\centering
	\includegraphics[width=\linewidth]{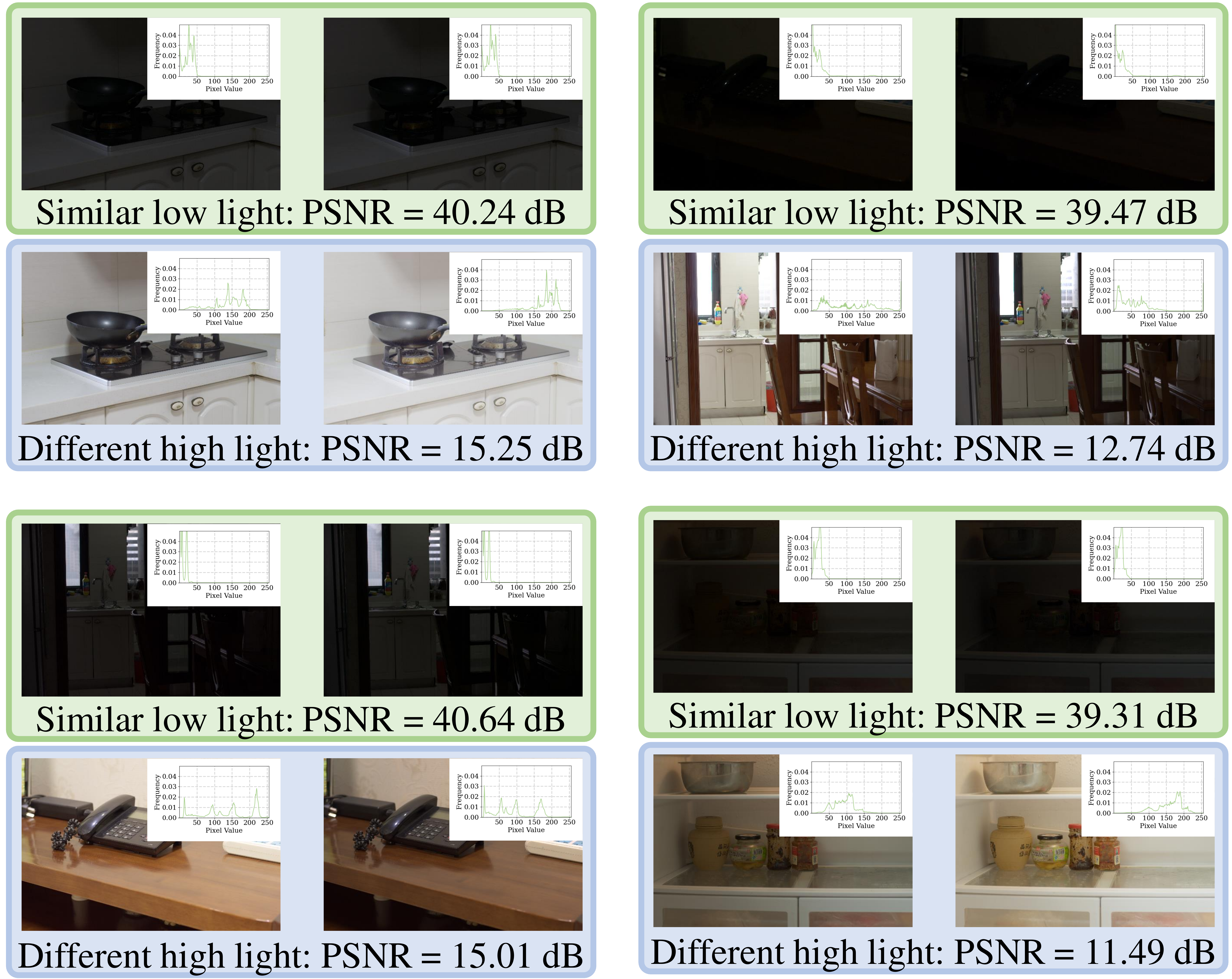} 
	%	\vspace{-1em}
	%	\caption{\textbf{Dynamic convolution.} 
		%		}
	\caption{\textbf{The similar low light images and corresponding different high light images.} 
		The training images do not even satisfy the same distribution, as similar low-light images are likely to be mapped to different high-light image. Yet LLIE methods typically are composed of static parameters. This forces the model to learn a general mapping, only well-suited for a part of images but become maladaptive and even harmful when facing other particular images.
	}	
	\label{fig:mapping}
	\vspace{-0.5em}
\end{figure}

%% file: Figure/PDE.tex
\begin{figure}
	\centering
	\includegraphics[width=\linewidth]{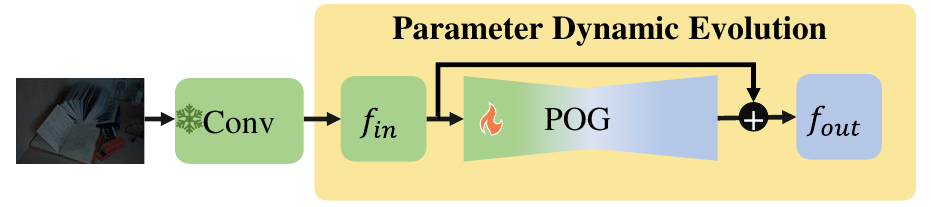} 
	\caption{\textbf{Overview of Parameter Dynamic Evolution (PDE).} 
%		PDE leverages POG to dynamic evolve appropriate feature from the original feature.
		PDE employs a parameter orthogonal generation technique and the corresponding generated parameters to simulate gene recombination and gene mutation, separately.
	}
	\label{fig:PDE}
	%	\vspace{-1em}
\end{figure}

%% file: Figure/POG.tex
\begin{figure*}
	\centering
%	\vspace{1em}
	\includegraphics[width=\linewidth]{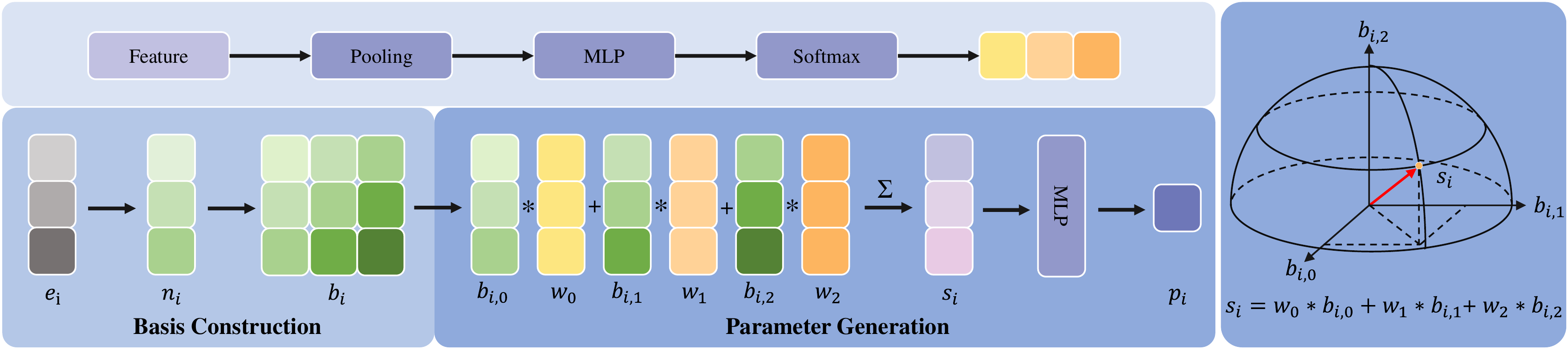} 
	\caption{\textbf{Overview of Parameter Orthogonal Generation (POG).} POG learns parameter embedding for each parameter, then constructs the orthogonal basis embeddings for the parameter, and finally generates specific parameters for the input image.}
	%	\vspace{0.5em}
	\label{fig:basis_conv}
\end{figure*}

%% file: 3_method.tex
\section{Method}

\subsection{Parameter Dynamic Evolution (PDE)}
\label{sec:PDE}

In this section, we describe the framework of parameter dynamic evolution (PDE) method. PDE evolves appropriate features from the original feature to adapt different images, as illustrated in Figure~\ref{fig:PDE}.

PDE is designed as a plug-and-play module that preserves the overall architecture and processing flow of the original neural network. The input feature \( f_{in} \) is processed by a dynamic block whose parameters are generated by POG, which will be discussed in Section~\ref{sec:POG}. The modified feature is then added to the original feature, producing the output feature \( f_{out} \).  

The dynamic block follows a bottleneck structure~\cite{efficient-bottleneck}, consisting of two convolutional layers that generate the output feature \( f_{out} \): 
\begin{equation}
	f_{out} =f_{in} + \mathcal{P}_{\theta_2}\circledast (\mathcal{P}_{\theta_1} \circledast f_{in} )
\end{equation}
where $\mathcal{P}_{\theta_1} \in \mathbb{R}^{D_c \times D_m \times D_k^2}$ and $\mathcal{P}_{\theta_2} \in \mathbb{R}^{D_m \times D_c \times D_k^2}$
are the dynamically generated parameters from POG. In addition, $\circledast$ denotes the convolution operation, $D_m$ denotes the channel dimension of the output of the first convolution, and $D_k$ denotes the kernel size. 
The constraint $D_m<D_c$ leads to a squeeze-and-excitation effect on the channel dimension, forming a bottleneck structure. 
The bottleneck structure significantly reduces the parameters~\cite{efficient-bottleneck} while aiding in the excitation of important information, according to information bottleneck theory~\cite{information-bottleneck}.

In summary, the output feature \( f_{out} \) evolved by PDE introduces greater adaptability and sensitivity to handle diverse images, thereby mitigating the gene effect.

\subsection{Parameter Orthogonal Generation (POG)}
\label{sec:POG}
 
In this section, we introduce parameter orthogonal generation (POG) technique, as illustrated in Figure~\ref{fig:basis_conv}.

Given an input image features $f_{in}$, POG generates specific parameters $\mathcal{P} \in \mathbb{R}^{C_{in} \times C_{out} \times D_k^2}$, where $C_{in}$, $C_{out}$, and $D_k$ denote the numbers of input channels, output channels, and convolution kernel size, respectively. POG comprises two primary steps: basis construction and parameter generation. 
Firstly, POG learns an embedding for each parameter and constructs orthogonal basis embeddings through the basis generation process. 
Subsequently, POG adaptively weights the basis embeddings to generate the specific embedding for the specific image and decodes specific parameters from the specific embedding. 

%\paragraph{Basis construction.} 
\noindent\textbf{Basis construction.}
Initially, POG learns parameter embeddings $\mathcal{E}_p \in \mathbb{R}^{N \times D_e \times 1}$ for the parameters $P$, where $N=C_{in} \times C_{out} \times D_k^2$ and $D_e$ represents the embedding dimension. These embeddings, denoted as $\mathcal{E}_p =[ e_{1},e_2,\cdots,e_{N}]$, correspond to each parameter $e_{i}$ individually. 
After that, POG normalizes each column vector embedding $e_{i}$ to obtain the normalized embeddings $\mathcal{N}_{p}$.

Next, POG constructs basis embeddings $\mathcal{B}_p$ for parameters based on the normalized embeddings $\mathcal{N}_{p}$:
\begin{equation}
	\mathcal{B}_{p} = I - 2\mathcal{N}_{p}\mathcal{N}_{p}^T .
	\label{eq:basis}
\end{equation}
where $I$ is the identity matrix. 
Here, $\mathcal{B}_{p} \in \mathbb{R}^{N \times D_e \times D_e}$ consists of basis embeddings $b_{i} \in \mathbb{R}^{D_e \times D_e}$. Each basis embeddings $b_{i}$ consist of one set of $D_e $ orthogonal bases for each parameter $e_{i}$~\cite{Matrix_analysis}, where $b_{i,j} \in \mathbb{R}^{D_e \times 1}, 1 \leq j \leq D_e$. Further theory guarantee regarding orthogonal bases is provided in the supplementary materials. The basis embeddings $\mathcal{B}_{p}$ are fixed after training.

\noindent\textbf{Parameter Generation.}
The specific parameters for each image are decoded from specific embeddings, which are constructed by adaptively weighting the basis embeddings.

The weights, derived from the input $f_{in}$, are obtained through the following process. 
Firstly, POG averages the spatial space of input $f_{in}$, then passes them through a 2-layer MLP~\cite{MLP}, and finally applies Softmax to obtain the weights $\mathcal{W}=[w_1, w_2,\cdots, w_{D_e}]^T \in \mathbb{R}^{D_e \times 1}$:
\begin{equation}
	\mathcal{W} = \text{Softmax}(\mathcal{M}_{\theta_3}( \text{Pooling}(f_{in}) )).
\end{equation}
where $\mathcal{M}_{\theta_3}$ is a 2-layer MLP parameterized by $\theta_3$.

For each parameter, POG adaptively weights the basis embeddings to derive the specific embedding $\mathcal{S}_p = [s_{1},s_{2},\cdots,s_{N} ] \in \mathbb{R}^{N \times D_e}$ specialized for the input $f_{in}$:
\begin{equation}
	s_{i} = \sum_{j=1}^{D_e} w_j b_{i,j}.
\end{equation}
This specific parameter embedding $\mathcal{S}_p$ is then decoded using a 2-layer MLP $\mathcal{M}_{\theta_4}$ parameterized by $\theta_4$, extracting the final parameters $\mathcal{P}$:
\begin{equation}
	\mathcal{P} = \mathcal{M}_{\theta_4}(\mathcal{S}_p),
\end{equation}
The MLP $\mathcal{M}_{\theta_4}$ decodes a parameter from the corresponding specific embedding $s_{i}$. After reshaping the shape of parameters $\mathcal{P}$, the generation process is concluded.

In summary, POG learns orthogonal basis embeddings for single parameters, thus avoiding the correlation in the embeddings and preventing the excessive expression of similar parameters.

%% file: 4_experiment.tex
\section{Experiments}

\input{Table/tab_detect_and_reduce}

\label{sec:Experiments}
\subsection{Implementation Details}

In the experiments, the channel dimension \(D_m\) is typically set to 32, while the embedding dimension \(D_e\) is set to 64. PDE is inserted after the attention mechanism in the decoder of the UNet-like architecture because we find that attention mechanism exhibit more gene effects than in other mechanisms. Specifically, the query (\(Q\)), key (\(K\)), and value (\(V\)) of the attention mechanism are concatenated and then fed into PDE. 
Our method requires 10k fine-tuning steps, while the original training required 320k steps.
Additional dataset details, implementation specifics, and visual results are provided in the supplementary materials.

\subsection{Gene Effect}

\input{Table/compare_with_other_MR_methods}

\input{Table/tab_com_llie}
\input{Figure/fig_com_llie}

\input{Figure/fig_com_llie_unpair}
\input{Table/tab_ab_2.tex}

\input{Table/tab_com_llie_unpair}

In this section, we detect the observed gene effect across different methods, evaluate the capability of our method in handling the gene effect and evaluate the capability of other possible methods in handling the gene effect.

Here, we use the method in motivation experiments and observations (Section~\ref{sec:motivation_experiments}) to {\bf{d}}etect {\bf{g}}ene  {\bf{e}}ffect, denoted this metric as {\bf{DGE}}.
Specifically, for each layer to be detected, we reset the parameters to random values and then calculate the differences between the images enhanced by the original model and those enhanced by the model with the reset parameters. 
Similar to PSNR, we use the logarithmic MSE to represent the differences, and the average logarithmic MSE across all layers is defined as DGE.
For a given trained model \(F\), let \(F_i\) represent the model where parameters of the \(i\)-th layer to be detected are reset. For a set of \(m\) test images \(x_j\), the metric DGE, that presents the gene effect level, is calculated as:

\begin{equation}
	\text{DGE} = \sum_{i=1}^{n} \sum_{j=1}^{m} 
	10 \cdot \log_{10} \left( \frac{I_{\text{max}}^2}{\text{MSE}(F(x_j), F_i(x_j))} \right) ,
\end{equation}
where \( I_{\text{max}} \) is the maximum pixel value of the image, typically 255 for 8-bit images.

A larger value of DGE suggests that even when the parameters are reset to random values, the model still produces results similar to a well-trained model with small differences, indicating more gene effects. Conversely, a smaller DGE implies weaker gene effects.

Firstly, we detect the gene effect across different methods. As shown in Table~\ref{tab:DGE}, different methods all exhibit gene effects.

Next, we also evaluate the capability of our method in handling the gene effect on different architectures. As shown in Table~\ref{tab:DGE}, our method significantly reduce the DGE, effectively decreasing the gene effect. 

Finally, we also evaluate the effectiveness of other potential methods in mitigating gene effects. Specifically, we mainly focus on pruning methods, which directly remove parameters that exhibit gene effects.
However, as shown in Table \ref{tab:DGE_PSNR}, these methods not only fail to effectively reduce gene effects but also degrade enhancement performance.
On the LOL-v2-synthetic~\cite{LOLv2} dataset, many methods (IENNP~\cite{IENNP} and FPGM~\cite{FPGM}) reduce 10$\%$ channels, but the LLIE performance collapses directly.
This phenomenon may result from the shift in the parameter distribution~\cite{DFQ}, leading to color distortion in the output images. Such distortion may be acceptable for high-level tasks but is unacceptable for LLIE, which aims to enhance color and illumination.

\subsection{Low-Light Image Enhancement}

In this section, we evaluate the enhancement performance. Our method achieves varying degrees of PSNR improvement based on different gene effects.

For paired datasets, we conduct experiments following previous research~\cite{retinexmamba,CIDNet}, evaluating our method on the popular LOL-v1~\cite{RetinexNet}, LOL-v2-real~\cite{LOLv2}, and LOL-v2-synthetic~\cite{LOLv2} datasets.
Table \ref{tab:sota_llie} presents a quantitative comparison of various methods.
Our method achieves varying degrees of PSNR improvement based on different gene effects.
Specifically, our method achieves a PSNR improvement of about 1 dB compared to the Restormer method and over 0.3 dB compared to the CIDNet method. 
Figure \ref{fig:llie} illustrates the visual results for the LOL datasets, demonstrating our method adapts to different images and learns accurate color. 

For unpaired datasets, our method also achieves effective improvement, as shown in Table~\ref{tab:unpaired} and Figure~\ref{fig:unpair}. More experiments and visual results are provided in the supplementary materials.

\subsection{Ablation Study}

%In this section, we conduct a comprehensive ablation study of our method.

\noindent\textbf{Component Analysis.}
In the ``Restormer+PDE" setting, the parameters are generated by traditional dynamic convolutions and in ``Restormer+PDE+POG", the parameters are generated by POG.
Incorporating only static convolutions into Restormer results in a slight improvement as demonstrated in Table \ref{table:ab-com}. Both our PDE and POG improve the performance, highlighting the effectiveness of each technique.

\noindent\textbf{Hyperparameter Analysis.}
We further investigate the impact of different hyperparameter settings, as shown in Table \ref{table:ab-arf} and Table \ref{table:ab-bpg}.
Increasing the dimension $D_m$ leads to an increase in FLOPs but slightly improves the PSNR.
This result aligns with the design goal of employing the bottleneck structure in PDE.
In addition, the computation is more sensitive to the dimension $D_m$ than the embedding dim $D_e$. Thus, for bigger methods (such as Restormer), we set smaller dimensions $D_m$ (usually $D_m=4$) to accelerate computation.

%% file: Table/tab_detect_and_reduce.tex
\begin{table}[t]
	\begin{center}
		%\vspace{-1mm}
		\small
		\setlength{\tabcolsep}{4.7pt}
		\caption{{\bf Gene effect among attention mechanism.}
		The DGE (↓) metric is employed to detect gene effect. 
		A smaller DGE indicates a greater difference in the output results before and after resetting, reflecting lower gene effect.
		}
				  
		\vspace{-0.5em}
		\begin{tabular}{l | c c c}
			\toprule
			Methods
			&LOL-v1
			&LOL-v2-real
			&LOL-v2-syn\\
			\midrule
			SNR-Net~\cite{SNRNet}               & 40.79& 41.54& 36.63 \\ 
			LLformer~\cite{LLformer}            & 49.94& -& - \\ 
			Retinexmamba~\cite{retinexmamba}    & 42.86& 40.76& 42.64 \\
			\midrule
			Restormer~\cite{Restormer}          & 48.94& 47.34 & 49.53 \\
			\textbf{Restormer+Ours}             & 45.09& 45.09 & 47.78 \\
			\midrule
			Retinexformer~\cite{Retinexformer}  & 34.88& 36.78
												& \textcolor{black}{36.35} \\
			\textbf{Retinexformer+Ours}         & \textcolor{black}{29.03}
												& \textcolor{black}{33.42}
												& \textcolor{black}{34.86} \\
			\midrule
			CIDNet~\cite{CIDNet}                & 33.46& 33.57& 37.02 \\
			\textbf{CIDNet+Ours}                & \textcolor{black}{33.40}
												& \textcolor{black}{31.99}
												& 36.58 \\
			
			\bottomrule
		\end{tabular}
		\label{tab:DGE}
	\end{center}
	\vspace{-1em}
\end{table}

%% file: Table/compare_with_other_MR_methods.tex
\begin{table}[t]
	\begin{center}
		\small
		\setlength{\tabcolsep}{5.8pt}
		\caption{
			{Evaluation (PSNR ↑ / DGE ↓) of other possible methods in handling gene effects.}	
		}
		\vspace{-0.5em}
		
\scalebox{1.0}{		
	\begin{tabular}{l | c c c}
			\toprule
			
			&LOL-v1
			&LOL-v2-real
			&LOL-v2-syn\\
			\cline{2-4}
			
			\multirow{-2}{*}{Methods}
			
			& PSNR / DGE
			& PSNR / DGE
			& PSNR / DGE
%			& PSNR↑/DGE↓
%			& PSNR↑/DGE↓
%			& PSNR↑/DGE↓
			\\
			\midrule
Restormer                   & 20.91 / 48.94   & 20.79 / 47.34      & 24.06 / 49.53 \\
\midrule
PC-0.1~\cite{pruning} & 20.94 / 49.21   & 20.73 / 46.21 & 24.00 / 48.11  \\
PC-0.2~\cite{pruning} & 20.83 / 50.00   & 20.99 / 45.63 & 23.82 / 48.19 \\
PC-0.3~\cite{pruning} & 20.66 / 51.55   & 11.00 / {-------} & {12.63 / -------}  \\
IENNP~\cite{IENNP}     &18.52 / 49.21 & 18.79 / 46.26 & {13.98 / -------}   \\

FPGM~\cite{FPGM}   &18.07 / 48.89  & 20.66 / 46.07 & 13.31 / {-------} \\ %45.83 \\
ZeroQ~\cite{ZeroQ}          & { 7.98} / {-------}  & { 9.75} / {-------}& { 9.81} / {-------}\\
\midrule 
\textbf{Ours} & 21.88 / 45.09   & 21.49 / 45.09 & 24.56 / 47.78  \\

			\bottomrule
		\end{tabular}}
		\label{tab:DGE_PSNR}
			\vspace{-1em}
	\end{center}
\end{table}

%% file: Table/tab_com_llie.tex
%\vspace{-1em}
\begin{table*}[!t]
	\begin{center}
	%\vspace{-1mm}
	\small
	\setlength{\tabcolsep}{8.5pt}
%	\caption{Comparison on low-light image enhancement benchmarks.}		  
	\caption{Quantitative comparison (PSNR $\uparrow$ and SSIM $\uparrow$) on paired datasets.
	Our techniques improve LLIE performance.
	}
		\vspace{-0.5em}
		\begin{tabular}{l | c| c| c c | c c | c c }
		\toprule
		{}
		&{}
		&{}
		&\multicolumn{2}{c|}{LOL-v1~\cite{RetinexNet}}
		&\multicolumn{2}{c|}{LOL-v2-real~\cite{LOLv2}} 
		&\multicolumn{2}{c}{LOL-v2-syn~\cite{LOLv2}}\\
		\multirow{-2}{*}{Methods} & 
%		\multirow{-2}{*}{Venue} & 
		\multirow{-2}{*}{Publication} & 
		\multirow{-2}{*}{FLOPs (G)} & 
		PSNR~$\textcolor{black}{\uparrow}$ & SSIM~$\textcolor{black}{\uparrow}$ & PSNR~$\textcolor{black}{\uparrow}$ & SSIM~$\textcolor{black}{\uparrow}$ & PSNR~$\textcolor{black}{\uparrow}$ & SSIM~$\textcolor{black}{\uparrow}$ \\
		\midrule

		RetinexNet~\cite{RetinexNet} &BMVC 2018 &  587.47 
		& 16.77 & 0.560 & 15.47 & 0.567  &17.13  &0.798 \\
		KinD~\cite{Kind}  & MM 2019&   34.99
		&20.86 &0.790  &14.74 &0.641  &13.29 &0.578    \\
		Enlightengan~\cite{Enlightengan}& TIP 2021 &   61.01  
		&17.48  &0.650    &18.23  &0.617   &16.57   &0.734   \\
		RUAS~\cite{Kind}  & CVPR 2021&   0.83
		&18.23 &0.720  &18.37 &0.723  &16.55 &0.652    \\ 
		SNRNet~\cite{SNRNet} & CVPR2022 &   26.35
		&24.61 &0.842&21.48&\textcolor{black}{0.849}&24.14 &0.928  \\ 
		LLformer~\cite{LLformer} & AAAI 2023 &  22.52
		&23.65 &0.816&20.06& 0.792&24.04 &0.909  \\ 
		
		GSAD~\cite{GSAD}&NeurIPS 2023 &-
		&22.88&0.849 &20.19&0.847 &24.22 &0.927
		\\
		QuadPrior~\cite{QuadPrior} &CVPR 2024 &-
		& 20.31&0.808 &-&- &- &-
		\\
		RSFNet~\cite{rsfnet}&CVPR 2024 &-
		&  19.39&0.755 &19.27&0.738 &- &-
		\\
		Retinexmamba~\cite{retinexmamba}& Arxiv 2024 & 42.82   
		& 24.03& 0.827& 22.45& {0.844}& \textcolor{black}{25.89} &\textcolor{black}{0.935} \\
		
		\midrule
		Restormer~\cite{Restormer}& CVPR 2022 &   144.25 
		&20.91        &0.788        &20.79         &0.816         &24.06     &0.919        \\
		
		\textbf{Restormer+Ours} & - &   145.99 
		&21.88&0.797
		&21.49&0.813
		&24.56&0.926
		\\
		\midrule
		Retinexformer~\cite{Retinexformer}& ICCV 2023 & 15.85 &\textcolor{black}{25.16}&0.845&22.80&0.840&25.67 &\textcolor{black}{0.930}\\
		
		\textbf{Retinexformer+Ours}&- & 16.56 
		&\textcolor{black}{25.29}&0.845
		&22.87&0.842 &\textcolor{black}{25.78} &\textcolor{black}{0.930} \\
		\midrule
		
		CIDNet~\cite{CIDNet}& Arxiv 2024 &  7.57
		&23.81&\textcolor{black}{0.857} &\textcolor{black}{23.90}&\textcolor{black}{0.866}    &25.71 &\textcolor{black}{0.942}\\
		
		\textbf{CIDNet+Ours} & - &  8.17
		&23.97&\textcolor{black}{0.859}
		&\textcolor{black}{24.21}&\textcolor{black}{0.866}&\textcolor{black}{26.02}&\textcolor{black}{0.942} \\
		
		\bottomrule
		\end{tabular}
	\label{tab:sota_llie}
	\end{center}
\end{table*}

%% file: Figure/fig_com_llie.tex
\begin{figure*}
	\centering
%	\vspace{-1em}
	\includegraphics[width=\linewidth]{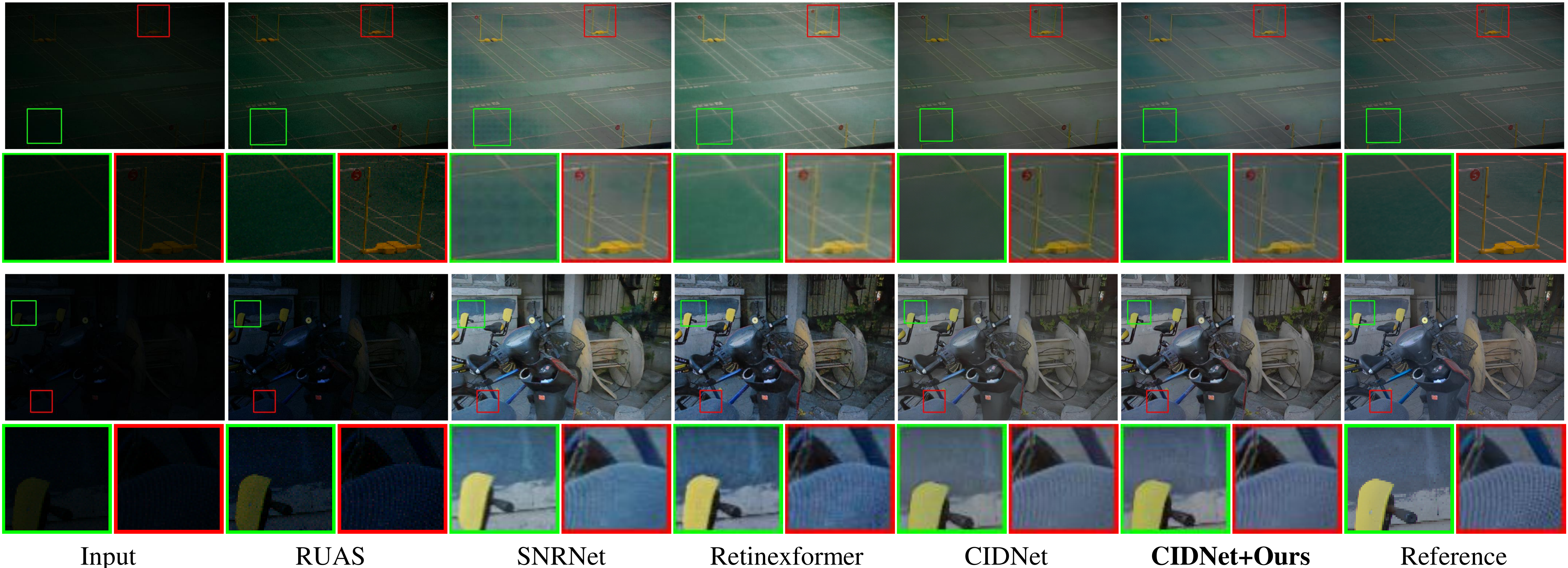} 
	\caption{Qualitative comparison on LOL-v1~\cite{RetinexNet} and LOL-v2~\cite{LOLv2} datasets.
%	The images ehnhanced by our method present more vibrant colors.
	}
	\label{fig:llie}
%	\vspace{-1em}
\end{figure*}

%% file: Figure/fig_com_llie_unpair.tex
\begin{figure*}
	\centering
	\includegraphics[width=\linewidth]{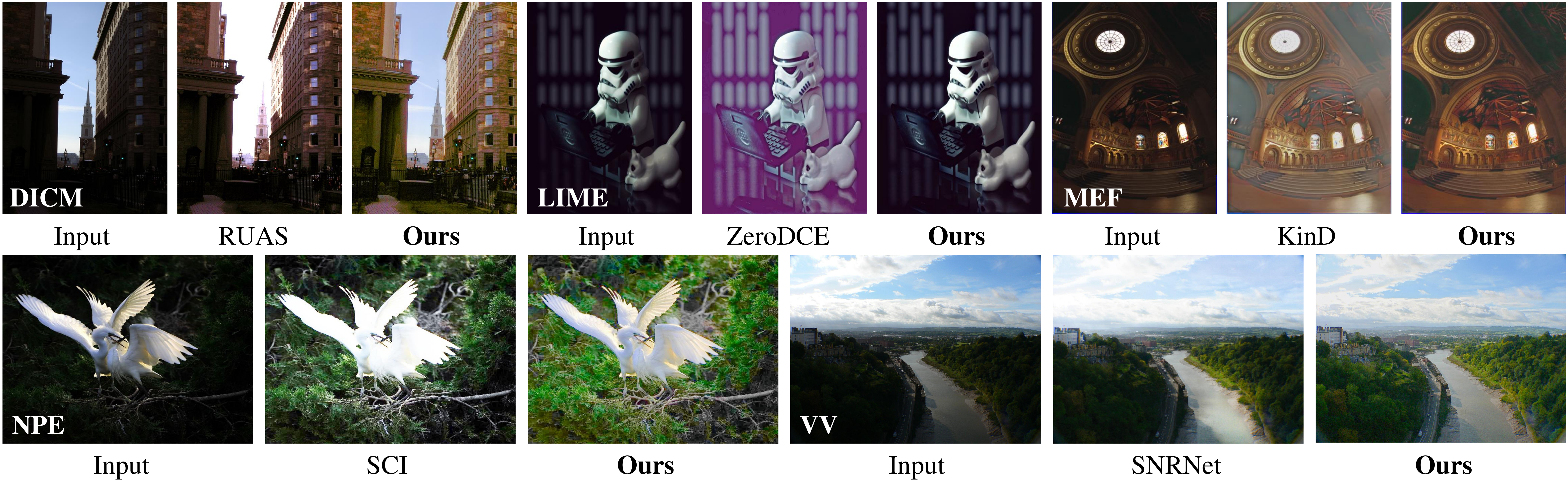} 
	\caption{Qualitative comparison on DICM~\cite{DICM}, LIME~\cite{LIME}, MEF~\cite{MEF}, NPE~\cite{NPE}, and VV~\cite{VV} datasets. }
	\label{fig:unpair}
\end{figure*}

%% file: Table/tab_ab_2.tex
\begin{table*}[tbp]
	%\begin{table*}[htbp]
	\parbox{.39\textwidth}{
		\centering
		\small
		\captionof{table}{Ablation study on the PDE technique and POG technique.}
		\vspace*{-0.5em}
		\label{table:ab-com}
		%		\vspace{-2mm}
		\setlength{\tabcolsep}{4pt}
		\scalebox{1}{
	\begin{tabular}{l|c c}
		\toprule
		Methods & PSNR ↑ &FLOPs (G) ↓\\
		\midrule
		Restormer & 20.91 & 144.25 \\
		Restormer + static conv & 21.18 & 145.69\\
		\midrule
		Restormer + PDE & 21.60 & 145.87 \\
		Restormer + PDE + POG & 21.88 &145.99 \\
		\bottomrule
	\end{tabular}
	}
	}
	\hfill
	\parbox{.28\textwidth}{
		\centering
		\small
		\captionof{table}{Ablation study on hyperparameters of PDE.}
		\vspace*{-0.5em}
		\label{table:ab-arf}
		%		\vspace{-2mm}
		\setlength{\tabcolsep}{4pt}
		\scalebox{1}{
			\begin{tabular}{l|c c}
				\toprule
				Methods & PSNR ↑ &FLOPs (G) ↓\\
				\midrule
				Restormer & 20.91 &144.25\\
				\midrule
				$D_m=4$  & 21.88 &145.99 \\
				$D_m=8$  & 21.92 &147.55\\
				$D_m=16$ & 21.77 &150.67 \\
				\bottomrule
		\end{tabular}}
	}
	\hfill
	\parbox{.28\textwidth}{
		\centering
		\captionof{table}{Ablation study on hyperparameters of POG.}
		\small
		\vspace*{-0.5em}
		\label{table:ab-bpg}
		%		\vspace{-2mm}
		\setlength{\tabcolsep}{4pt}
		\scalebox{1}{
			\begin{tabular}{l|c c}
				\toprule
				Methods & PSNR ↑ &FLOPs (G) ↓\\
				\midrule
				Restormer & 20.91 &144.25\\
				\midrule
				$D_e=16$ & 21.53&145.88 \\
				$D_e=32$ & 21.86&145.90\\
				$D_e=64$ & 21.88&145.99 \\
				\bottomrule
		\end{tabular}}
	}
\end{table*}

%% file: Table/tab_com_llie_unpair.tex
\begin{table}[tbp]
	\centering
	\small
	\vspace*{1em}
	\setlength{\tabcolsep}{3pt}
	\caption{Quantitative comparison (NIQE $\downarrow$) on unpaired datasets.}
		\begin{tabular}{l|ccccc c}
			\toprule
{Methods}&{DICM}& {LIME} & {MEF} & {NPE} &{VV} & Mean\\
			\midrule
			KinD \cite{Kind}&
			5.15& 	
			5.03& 	
			5.47& 	
			4.98& 	
			4.30&
			4.99
			\\
			ZeroDCE \cite{ZeroDCE}&
			4.58&
			5.82&
			4.93&
			4.53&
			4.81&
			4.93
			\\
			
			RUAS \cite{RUAS}&
			5.21& 	
			4.26& 	
			3.83& 	
			5.53& 	
			4.29&
			4.62
			\\
			LLFlow \cite{LLFlow}&
			4.06& 
			4.59& 
			4.70& 
			4.67& 
			4.04&
			4.41
			\\
			SNRNet \cite{SNRNet}&
			4.71& 
			5.74& 	
			4.18& 	
			4.32& 	
			9.87&
			5.76
			\\
			PairLIE \cite{PairLIE}&
			4.03& 	
			4.58& 	
			4.06& 	
			4.18& 	
			3.57&
			4.08
			
			\\
			
			GLARE \cite{GLARE}&
			3.61& 	
			4.52& 	
			3.66& 	
			4.19& 	
			-& 	
			4.10
			\\
			\midrule
			{Restormer}~\cite{Restormer}&
			3.49 & 4.31 & 3.71 & 3.97 & 2.93 & 3.68
			\\
			\textbf{Restormer+Ours} &
			3.42 & 4.25 & 3.66 & 3.96 & 2.81 & 3.62
			\\
			\midrule
			{Retinexformer}~\cite{Retinexformer}&
			3.85 &
			4.31 &
			3.67 &
			3.76 &
			3.09 &
			3.74
			\\
			\textbf{Retinexformer+Ours}&
		   3.51&4.00&3.62&3.92&3.00&3.61
			\\
			\midrule
			{CIDNet}~\cite{CIDNet}&
			{3.79}&	
			{4.13}&
			{3.56}&
			{3.74}&	
		    {3.21}&
		    3.67
			\\
			\textbf {CIDNet+Ours} &	
			3.50 &
			3.41 &
			3.08 &
			4.23 &
			3.19 &
			3.48 
			\\	
			
			\bottomrule
		\end{tabular}
		%	}
%	\vspace{-0.5em}
	\label{tab:unpaired}
\end{table}%

%% file: 6_conclusion.tex
\section{Conclusion}
In this paper, we observe and identify the counterintuitive gene effect in LLIE.
Inspired by biological gene evolution, we attribute the gene effect to static parameters.
To address the gene effect, we propose the parameter dynamic evolution to simulate gene dynamic evolution and mitigate the gene effect.
In the future, we will employ the gene effect to guide the effective architecture design.

%% file: main.bbl
\begin{thebibliography}{73}
\providecommand{\natexlab}[1]{#1}
\providecommand{\url}[1]{\texttt{#1}}
\expandafter\ifx\csname urlstyle\endcsname\relax
  \providecommand{\doi}[1]{doi: #1}\else
  \providecommand{\doi}{doi: \begingroup \urlstyle{rm}\Url}\fi

\bibitem[Bai et~al.(2024)Bai, Yin, and He]{retinexmamba}
Jiesong Bai, Yuhao Yin, and Qiyuan He.
\newblock Retinexmamba: Retinex-based mamba for low-light image enhancement.
\newblock \emph{arXiv preprint arXiv:2405.03349}, pages 1--15, 2024.

\bibitem[Cai et~al.(2020)Cai, Yao, Dong, Gholami, Mahoney, and Keutzer]{ZeroQ}
Yaohui Cai, Zhewei Yao, Zhen Dong, Amir Gholami, Michael~W Mahoney, and Kurt
  Keutzer.
\newblock Zeroq: A novel zero shot quantization framework.
\newblock \emph{Proceedings of the IEEE/CVF Conference on Computer Vision and
  Pattern Recognition (CVPR)}, pages 13169--13178, 2020.

\bibitem[Cai et~al.(2023)Cai, Bian, Lin, Wang, Timofte, and
  Zhang]{Retinexformer}
Yuanhao Cai, Hao Bian, Jing Lin, Haoqian Wang, Radu Timofte, and Yulun Zhang.
\newblock Retinexformer: One-stage retinex-based transformer for low-light
  image enhancement.
\newblock \emph{Proceedings of the IEEE/CVF International Conference on
  Computer Vision (ICCV)}, pages 12504--12513, 2023.

\bibitem[Casper et~al.(2019)Casper, Boix, D'Amario, Rodriguez, Guo, Vinken, and
  Kreiman]{Robustness}
Stephen Casper, Xavier Boix, Vanessa D'Amario, Christopher Rodriguez, Ling Guo,
  Kasper Vinken, and Gabriel Kreiman.
\newblock Removable and/or repeated units emerge in overparametrized deep
  neural networks.
\newblock \emph{arXiv preprint arXiv:1912.04783}, pages 1--9, 2019.

\bibitem[Chen et~al.(2020)Chen, Dai, Liu, Chen, Yuan, and
  Liu]{dynamicconvolution}
Yinpeng Chen, Xiyang Dai, Mengchen Liu, Dongdong Chen, Lu Yuan, and Zicheng
  Liu.
\newblock Dynamic convolution: Attention over convolution kernels.
\newblock \emph{Proceedings of the IEEE/CVF Conference on Computer Vision and
  Pattern Recognition (CVPR)}, pages 11030--11039, 2020.

\bibitem[Cheng and Shi(2004)]{HE3}
Hengda Cheng and XJ Shi.
\newblock A simple and effective histogram equalization approach to image
  enhancement.
\newblock \emph{Digital Signal Processing}, pages 158--170, 2004.

\bibitem[Darwin(1859)]{Darwin}
Charles Darwin.
\newblock Origin of the species.
\newblock 1859.

\bibitem[Dobzhansky(1951)]{Dobzhansky}
Theodosius Dobzhansky.
\newblock Genetics and the origin of species.
\newblock 1951.

\bibitem[Feng et~al.(2024)Feng, Zhang, Wang, Wu, Yan, and Zhang]{CIDNet}
Yixu Feng, Cheng Zhang, Pei Wang, Peng Wu, Qingsen Yan, and Yanning Zhang.
\newblock You only need one color space: An efficient network for low-light
  image enhancement.
\newblock \emph{arXiv preprint arXiv:2402.05809}, pages 1--9, 2024.

\bibitem[Frank(1958)]{MLP}
Rosenblatt Frank.
\newblock The perceptron: A probabilistic model for information storage and
  organization in the brain.
\newblock \emph{Psychological Review}, page 386, 1958.

\bibitem[Fu et~al.(2023)Fu, Yang, Tu, Huang, Ding, and Ma]{PairLIE}
Zhenqi Fu, Yan Yang, Xiaotong Tu, Yue Huang, Xinghao Ding, and Kai-Kuang Ma.
\newblock Learning a simple low-light image enhancer from paired low-light
  instances.
\newblock \emph{Proceedings of the IEEE/CVF Conference on Computer Vision and
  Pattern Recognition (CVPR)}, pages 22252--22261, 2023.

\bibitem[Guo et~al.(2024)Guo, Li, Dai, Ouyang, Ren, and Xia]{mambair}
Hang Guo, Jinmin Li, Tao Dai, Zhihao Ouyang, Xudong Ren, and Shu-Tao Xia.
\newblock Mambair: A simple baseline for image restoration with state-space
  model.
\newblock \emph{arXiv preprint arXiv:2402.15648}, pages 1--19, 2024.

\bibitem[Guo et~al.(2016)Guo, Li, and Ling]{LIME}
Xiaojie Guo, Yu Li, and Haibin Ling.
\newblock Lime: Low-light image enhancement via illumination map estimation.
\newblock \emph{IEEE Transactions on Image Processing (TIP)}, pages 982--993,
  2016.

\bibitem[Han et~al.(2015)Han, Pool, Tran, and Dally]{pruning}
Song Han, Jeff Pool, John Tran, and William Dally.
\newblock Learning both weights and connections for efficient neural network.
\newblock \emph{Advances in Neural Information Processing Systems}, pages 1--9,
  2015.

\bibitem[Hao et~al.(2017)Hao, Asim, Igor, Hanan, and HansPeter]{IENNP}
Li Hao, Kadav Asim, Durdanovic Igor, Samet Hanan, and Graf HansPeter.
\newblock Importance estimation for neural network pruning.
\newblock \emph{International Conference on Learning Representations (ICLR)},
  pages 1--12, 2017.

\bibitem[He et~al.(2019)He, Liu, Wang, Hu, and Yang]{FPGM}
Yang He, Ping Liu, Ziwei Wang, Zhilan Hu, and Yi Yang.
\newblock Filter pruning via geometric median for deep convolutional neural
  networks acceleration.
\newblock \emph{Proceedings of the IEEE Conference on Computer Vision and
  Pattern Recognition (CVPR)}, pages 4340--4349, 2019.

\bibitem[Horn and Johnson(1990)]{Matrix_analysis}
Roger~A Horn and Charles~R Johnson.
\newblock \emph{Matrix Analysis}.
\newblock Cambridge University Press, 1990.

\bibitem[Hou et~al.(2024)Hou, Zhu, Hou, Liu, Zeng, and Yuan]{GSAD}
Jinhui Hou, Zhiyu Zhu, Junhui Hou, Hui Liu, Huanqiang Zeng, and Hui Yuan.
\newblock Global structure-aware diffusion process for low-light image
  enhancement.
\newblock \emph{Advances in Neural Information Processing Systems}, pages
  79734--79747, 2024.

\bibitem[Hssayni et~al.(2022)Hssayni, Joudar, and
  Ettaouil]{hssayni2022redundancy}
El~Houssaine Hssayni, Nour-Eddine Joudar, and Mohamed Ettaouil.
\newblock Krr-cnn: kernels redundancy reduction in convolutional neural
  networks.
\newblock \emph{Neural Computing and Applications}, pages 2443--2454, 2022.

\bibitem[Huang et~al.(2012)Huang, Cheng, and Chiu]{GC1}
ShihChia Huang, FanChieh Cheng, and YiSheng Chiu.
\newblock Efficient contrast enhancement using adaptive gamma correction with
  weighting distribution.
\newblock \emph{IEEE Transactions on Image Processing (TIP)}, pages 1032--1041,
  2012.

\bibitem[Jiang et~al.(2023)Jiang, Luo, Fan, Han, and Liu]{DiffLL}
Hai Jiang, Ao Luo, Haoqiang Fan, Songchen Han, and Shuaicheng Liu.
\newblock Low-light image enhancement with wavelet-based diffusion models.
\newblock \emph{ACM Transactions on Graphics (TOG)}, pages 1--14, 2023.

\bibitem[Jiang et~al.(2021)Jiang, Gong, Liu, Cheng, Fang, Shen, Yang, Zhou, and
  Wang]{Enlightengan}
Yifan Jiang, Xinyu Gong, Ding Liu, Yu Cheng, Chen Fang, Xiaohui Shen, Jianchao
  Yang, Pan Zhou, and Zhangyang Wang.
\newblock Enlightengan: Deep light enhancement without paired supervision.
\newblock \emph{IEEE Transactions on Image Processing (TIP)}, pages 2340--2349,
  2021.

\bibitem[Jobson et~al.(1997{\natexlab{a}})Jobson, Rahman, and Woodell]{RT2}
Daniel~J Jobson, Zia-ur Rahman, and Glenn~A Woodell.
\newblock A multiscale retinex for bridging the gap between color images and
  the human observation of scenes.
\newblock \emph{IEEE Transactions on Image Processing (TIP)}, pages 965--976,
  1997{\natexlab{a}}.

\bibitem[Jobson et~al.(1997{\natexlab{b}})Jobson, Rahman, and Woodell]{RT3}
Daniel~J Jobson, Zia-ur Rahman, and Glenn~A Woodell.
\newblock Properties and performance of a center/surround retinex.
\newblock \emph{IEEE Transactions on Image Processing (TIP)}, pages 451--462,
  1997{\natexlab{b}}.

\bibitem[Kimura et~al.(1968)]{Kimura}
Motoo Kimura et~al.
\newblock Evolutionary rate at the molecular level.
\newblock \emph{Nature}, pages 624--626, 1968.

\bibitem[Lee et~al.(2013)Lee, Lee, and Kim]{DICM}
Chulwoo Lee, Chul Lee, and Chang-Su Kim.
\newblock Contrast enhancement based on layered difference representation of 2d
  histograms.
\newblock \emph{IEEE Transactions on Image Processing (TIP)}, pages 5372--5384,
  2013.

\bibitem[Li et~al.(2022{\natexlab{a}})Li, Guo, Han, Jiang, Cheng, Gu, and
  Loy]{LLIE_suervey}
Chongyi Li, Chunle Guo, Linghao Han, Jun Jiang, Mingming Cheng, Jinwei Gu, and
  Chen~Change Loy.
\newblock Low-light image and video enhancement using deep learning: A survey.
\newblock \emph{IEEE Transactions on Pattern Analysis and Machine Intelligence
  (TPAMI)}, pages 9396--9416, 2022{\natexlab{a}}.

\bibitem[Li et~al.(2022{\natexlab{b}})Li, Guo, and Loy]{ZeroDCE}
Chongyi Li, Chunle Guo, and Chen~Change Loy.
\newblock Learning to enhance low-light image via zero-reference deep curve
  estimation.
\newblock \emph{IEEE Transactions on Pattern Analysis and Machine Intelligence
  (TPAMI)}, pages 4225--4238, 2022{\natexlab{b}}.

\bibitem[Li et~al.(2018)Li, Liu, Yang, Sun, and Guo]{RT4}
Mading Li, Jiaying Liu, Wenhan Yang, Xiaoyan Sun, and Zongming Guo.
\newblock Structure-revealing low-light image enhancement via robust retinex
  model.
\newblock \emph{IEEE Transactions on Image Processing (TIP)}, pages 2828--2841,
  2018.

\bibitem[Liang et~al.(2021)Liang, Wang, Quan, Chen, Liu, Ling, and
  Xu]{Detection}
Jinxiu Liang, Jingwen Wang, Yuhui Quan, Tianyi Chen, Jiaying Liu, Haibin Ling,
  and Yong Xu.
\newblock Recurrent exposure generation for low-light face detection.
\newblock \emph{IEEE Transactions on Multimedia}, pages 1609--1621, 2021.

\bibitem[Liu et~al.(2021)Liu, Xu, Yang, Fan, and Huang]{Detection3}
Jiaying Liu, Dejia Xu, Wenhan Yang, Minhao Fan, and Haofeng Huang.
\newblock Benchmarking low-light image enhancement and beyond.
\newblock \emph{International Journal of Computer Vision}, pages 1153--1184,
  2021.

\bibitem[Ma et~al.(2015)Ma, Zeng, and Wang]{MEF}
Kede Ma, Kai Zeng, and Zhou Wang.
\newblock Perceptual quality assessment for multi-exposure image fusion.
\newblock \emph{IEEE Transactions on Image Processing (TIP)}, pages 3345--3356,
  2015.

\bibitem[Moran et~al.(2020)Moran, Marza, McDonagh, Parisot, and
  Slabaugh]{DeepLPF}
Sean Moran, Pierre Marza, Steven McDonagh, Sarah Parisot, and Gregory Slabaugh.
\newblock Deeplpf: Deep local parametric filters for image enhancement.
\newblock \emph{Proceedings of the IEEE/CVF Conference on Computer Vision and
  Pattern Recognition (CVPR)}, pages 12826--12835, 2020.

\bibitem[Morgan et~al.(1923)Morgan, Sturtevant, Muller, and Bridges]{Morgan}
Thomas~Hunt Morgan, Alfred~Henry Sturtevant, Hermann~Joseph Muller, and
  Calvin~Blackman Bridges.
\newblock The mechanism of mendelian heredity.
\newblock 1923.

\bibitem[Muller(1927)]{Muller}
Hermann~J Muller.
\newblock Artificial transmutation of the gene.
\newblock \emph{Science}, pages 84--87, 1927.

\bibitem[Nagel et~al.(2019)Nagel, Baalen, Blankevoort, and Welling]{DFQ}
Markus Nagel, Mart~van Baalen, Tijmen Blankevoort, and Max Welling.
\newblock Data-free quantization through weight equalization and bias
  correction.
\newblock \emph{Proceedings of the IEEE/CVF International Conference on
  Computer Vision (ICCV)}, pages 1325--1334, 2019.

\bibitem[Ochman et~al.(2000)Ochman, Lawrence, and Groisman]{Ochman}
Howard Ochman, Jeffrey~G Lawrence, and Eduardo~A Groisman.
\newblock Lateral gene transfer and the nature of bacterial innovation.
\newblock \emph{nature}, pages 299--304, 2000.

\bibitem[Ooi and Isa(2010)]{HE6}
Chen~Hee Ooi and Nor Ashidi~Mat Isa.
\newblock Quadrants dynamic histogram equalization for contrast enhancement.
\newblock \emph{IEEE Transactions on Consumer Electronics}, pages 2552--2559,
  2010.

\bibitem[Pisano et~al.(1998)Pisano, Zong, Hemminger, DeLuca, Johnston, Muller,
  Braeuning, and Pizer]{HE4}
Etta~D Pisano, Shuquan Zong, Bradley~M Hemminger, Marla DeLuca, R~Eugene
  Johnston, Keith Muller, M~Patricia Braeuning, and Stephen~M Pizer.
\newblock Contrast limited adaptive histogram equalization image processing to
  improve the detection of simulated spiculations in dense mammograms.
\newblock \emph{Journal of Digital imaging}, pages 193--200, 1998.

\bibitem[Pizer et~al.(1987)Pizer, Amburn, Austin, Cromartie, Geselowitz, Greer,
  ter Haar~Romeny, Zimmerman, and Zuiderveld]{HE5}
Stephen~M Pizer, E~Philip Amburn, John~D Austin, Robert Cromartie, Ari
  Geselowitz, Trey Greer, Bart ter Haar~Romeny, John~B Zimmerman, and Karel
  Zuiderveld.
\newblock Adaptive histogram equalization and its variations.
\newblock \emph{Computer Vision, Graphics, and Image Processing}, pages
  355--368, 1987.

\bibitem[Risheng et~al.(2021)Risheng, Long, Jiaao, Xin, and Zhongxuan]{RUAS}
Liu Risheng, Ma Long, Zhang Jiaao, Fan Xin, and Luo Zhongxuan.
\newblock Retinex-inspired unrolling with cooperative prior architecture search
  for low-light image enhancement.
\newblock \emph{Proceedings of the IEEE/CVF Conference on Computer Vision and
  Pattern Recognition (CVPR)}, 2021.

\bibitem[Saini and Narayanan(2024)]{rsfnet}
Saurabh Saini and P.~J. Narayanan.
\newblock Specularity factorization for low light enhancement.
\newblock \emph{Proceedings of the IEEE/CVF Conference on Computer Vision and
  Pattern Recognition (CVPR)}, pages 1--12, 2024.

\bibitem[Sharma and Tan(2021)]{CNN3}
Aashish Sharma and Robby~T Tan.
\newblock Nighttime visibility enhancement by increasing the dynamic range and
  suppression of light effects.
\newblock \emph{Proceedings of the IEEE/CVF Conference on Computer Vision and
  Pattern Recognition (CVPR)}, pages 11977--11986, 2021.

\bibitem[Sharma et~al.(2018)Sharma, Diba, Neven, Brown, Van~Gool, and
  Stiefelhagen]{dynamicie}
Vivek Sharma, Ali Diba, Davy Neven, Michael~S Brown, Luc Van~Gool, and Rainer
  Stiefelhagen.
\newblock Classification-driven dynamic image enhancement.
\newblock pages 4033--4041, 2018.

\bibitem[Shen et~al.(2023)Shen, Zhao, and Zhang]{ADFNet}
Hao Shen, Zhong-Qiu Zhao, and Wandi Zhang.
\newblock Adaptive dynamic filtering network for image denoising.
\newblock \emph{Proceedings of the AAAI Conference on Artificial Intelligence
  (AAAI)}, pages 2227--2235, 2023.

\bibitem[Shi et~al.(2020)Shi, Zhong, Yang, Yang, and Lin]{dynamicsr2}
Yukai Shi, Haoyu Zhong, Zhijing Yang, Xiaojun Yang, and Liang Lin.
\newblock Ddet: Dual-path dynamic enhancement network for real-world image
  super-resolution.
\newblock \emph{IEEE Signal Processing Letters}, pages 481--485, 2020.

\bibitem[Tian et~al.(2024)Tian, Zhang, Zhang, Yang, and Ju]{dynamicsr}
Chunwei Tian, Xuanyu Zhang, Qi Zhang, Mingming Yang, and Zhaojie Ju.
\newblock Image super-resolution via dynamic network.
\newblock \emph{CAAI Transactions on Intelligence Technology}, pages 837--849,
  2024.

\bibitem[Tishby and Zaslavsky(2015)]{information-bottleneck}
Naftali Tishby and Noga Zaslavsky.
\newblock Deep learning and the information bottleneck principle.
\newblock \emph{2015 IEEE Information Theory Workshop}, pages 1--5, 2015.

\bibitem[Vonikakis et~al.(2018)Vonikakis, Kouskouridas, and Gasteratos]{VV}
Vassilios Vonikakis, Rigas Kouskouridas, and Antonios Gasteratos.
\newblock On the evaluation of illumination compensation algorithms.
\newblock \emph{Multimedia Tools and Applications}, pages 1--21, 2018.

\bibitem[Wang et~al.(2019)Wang, Zhang, Fu, Shen, Zheng, and Jia]{DeepUPE}
Ruixing Wang, Qing Zhang, Chiwing Fu, Xiaoyong Shen, Weishi Zheng, and Jiaya
  Jia.
\newblock Underexposed photo enhancement using deep illumination estimation.
\newblock \emph{Proceedings of the IEEE/CVF Conference on Computer Vision and
  Pattern Recognition (CVPR)}, pages 6849--6857, 2019.

\bibitem[Wang et~al.(2013)Wang, Zheng, Hu, and Li]{NPE}
Shuhang Wang, Jin Zheng, Hai-Miao Hu, and Bo Li.
\newblock Naturalness preserved enhancement algorithm for non-uniform
  illumination images.
\newblock \emph{IEEE Transactions on Image Processing (TIP)}, pages 3538--3548,
  2013.

\bibitem[Wang et~al.(2023{\natexlab{a}})Wang, Zhang, Shen, Luo, Stenger, and
  Lu]{LLformer}
Tao Wang, Kaihao Zhang, Tianrun Shen, Wenhan Luo, Bjorn Stenger, and Tong Lu.
\newblock Ultra-high-definition low-light image enhancement: A benchmark and
  transformer-based method.
\newblock \emph{Proceedings of the AAAI Conference on Artificial Intelligence
  (AAAI)}, pages 2654--2662, 2023{\natexlab{a}}.

\bibitem[Wang et~al.(2018)Wang, Wei, Yang, and Liu]{CNN4}
Wenjing Wang, Chen Wei, Wenhan Yang, and Jiaying Liu.
\newblock Gladnet: Low-light enhancement network with global awareness.
\newblock \emph{IEEE International Conference on Automatic Face and Gesture
  Recognition}, pages 751--755, 2018.

\bibitem[Wang et~al.(2024)Wang, Yang, Fu, and Liu]{QuadPrior}
Wenjing Wang, Huan Yang, Jianlong Fu, and Jiaying Liu.
\newblock Zero-reference low-light enhancement via physical quadruple priors.
\newblock \emph{Proceedings of the IEEE/CVF Conference on Computer Vision and
  Pattern Recognition (CVPR)}, pages 26057--26066, 2024.

\bibitem[Wang et~al.(2022)Wang, Wan, Yang, Li, Chau, and Kot]{LLFlow}
Yufei Wang, Renjie Wan, Wenhan Yang, Haoliang Li, Lap-Pui Chau, and Alex Kot.
\newblock Low-light image enhancement with normalizing flow.
\newblock \emph{Proceedings of the AAAI Conference on Artificial Intelligence
  (AAAI)}, pages 2604--2612, 2022.

\bibitem[Wang et~al.(2023{\natexlab{b}})Wang, Liu, Liu, Xu, and Liu]{IACG}
Yinglong Wang, Zhen Liu, Jianzhuang Liu, Songcen Xu, and Shuaicheng Liu.
\newblock Low-light image enhancement with illumination-aware gamma correction
  and complete image modelling network.
\newblock \emph{Proceedings of the IEEE/CVF International Conference on
  Computer Vision (ICCV)}, pages 13128--13137, 2023{\natexlab{b}}.

\bibitem[Wang et~al.(2009)Wang, Liang, and Liu]{GC3}
Zhiguo Wang, Zhihu Liang, and Chunliang Liu.
\newblock A real-time image processor with combining dynamic contrast ratio
  enhancement and inverse gamma correction for pdp.
\newblock \emph{Displays}, pages 133--139, 2009.

\bibitem[Wei et~al.(2018)Wei, Wang, Yang, and Liu]{RetinexNet}
Chen Wei, Wenjing Wang, Wenhan Yang, and Jiaying Liu.
\newblock Deep retinex decomposition for low-light enhancement.
\newblock \emph{British Machine Vision Conference (BMVC)}, pages 1--12, 2018.

\bibitem[Weng et~al.(2024)Weng, Yan, Tai, Qian, Yang, and Li]{MambaLLIE}
Jiangwei Weng, Zhiqiang Yan, Ying Tai, Jianjun Qian, Jian Yang, and Jun Li.
\newblock Mamballie: Implicit retinex-aware low light enhancement with
  global-then-local state space.
\newblock \emph{arXiv preprint arXiv:2405.16105}, pages 1--12, 2024.

\bibitem[Xia et~al.(2023)Xia, Zhang, Wang, Wang, Wu, Tian, Yang, and
  Van~Gool]{xia2023diffir}
Bin Xia, Yulun Zhang, Shiyin Wang, Yitong Wang, Xinglong Wu, Yapeng Tian,
  Wenming Yang, and Luc Van~Gool.
\newblock Diffir: Efficient diffusion model for image restoration.
\newblock \emph{Proceedings of the IEEE/CVF Conference on Computer Vision and
  Pattern Recognition (CVPR)}, pages 13095--13105, 2023.

\bibitem[Xu et~al.(2022)Xu, Wang, Fu, and Jia]{SNRNet}
Xiaogang Xu, Ruixing Wang, Chiwing Fu, and Jiaya Jia.
\newblock Snr-aware low-light image enhancement.
\newblock \emph{Proceedings of the IEEE/CVF Conference on Computer Vision and
  Pattern Recognition (CVPR)}, pages 17714--17724, 2022.

\bibitem[Xu et~al.(2020)Xu, Tseng, Tseng, Kuo, and Tsai]{xu2020degradations}
Yusyuan Xu, Shouyao~Roy Tseng, Yu Tseng, Hsienkai Kuo, and Yi-Min Tsai.
\newblock Unified dynamic convolutional network for super-resolution with
  variational degradations.
\newblock \emph{Proceedings of the IEEE/CVF Conference on Computer Vision and
  Pattern Recognition (CVPR)}, pages 12496--12505, 2020.

\bibitem[Yang et~al.(2019)Yang, Bender, Le, and Ngiam]{condconvolution}
Brandon Yang, Gabriel Bender, Quoc~V Le, and Jiquan Ngiam.
\newblock Condconv: Conditionally parameterized convolutions for efficient
  inference.
\newblock \emph{Advances in Neural Information Processing Systems}, pages
  1--11, 2019.

\bibitem[Yang et~al.(2021)Yang, Wang, Huang, Wang, and Liu]{LOLv2}
Wenhan Yang, Wenjing Wang, Haofeng Huang, Shiqi Wang, and Jiaying Liu.
\newblock Sparse gradient regularized deep retinex network for robust low-light
  image enhancement.
\newblock \emph{IEEE Transactions on Image Processing (TIP)}, pages 2072--2086,
  2021.

\bibitem[Yi et~al.(2023)Yi, Xu, Zhang, Tang, and Ma]{Diffretinex}
Xunpeng Yi, Han Xu, Hao Zhang, Linfeng Tang, and Jiayi Ma.
\newblock Diff-retinex: Rethinking low-light image enhancement with a
  generative diffusion model.
\newblock \emph{Proceedings of the IEEE/CVF International Conference on
  Computer Vision (ICCV)}, pages 12302--12311, 2023.

\bibitem[Zamir et~al.(2022{\natexlab{a}})Zamir, Arora, Khan, Hayat, Khan, Yang,
  and Shao]{MIRNet}
Syed~Waqas Zamir, Aditya Arora, Salman Khan, Munawar Hayat, Fahad~Shahbaz Khan,
  MingHsuan Yang, and Ling Shao.
\newblock Learning enriched features for fast image restoration and
  enhancement.
\newblock \emph{IEEE Transactions on Pattern Analysis and Machine Intelligence
  (TPAMI)}, pages 1934--1948, 2022{\natexlab{a}}.

\bibitem[Zamir et~al.(2022{\natexlab{b}})Zamir, Arora, Khan, Hayat, Khan, and
  Yang]{Restormer}
Syed~Waqas Zamir, Aditya Arora, Salman Khan, Munawar Hayat, Fahad~Shahbaz Khan,
  and Ming-Hsuan Yang.
\newblock Restormer: Efficient transformer for high-resolution image
  restoration.
\newblock \emph{Proceedings of the IEEE/CVF Conference on Computer Vision and
  Pattern Recognition (CVPR)}, pages 5728--5739, 2022{\natexlab{b}}.

\bibitem[Zhang et~al.(2018)Zhang, Zuo, and Zhang]{zhang2018degradations}
Kai Zhang, Wangmeng Zuo, and Lei Zhang.
\newblock Learning a single convolutional super-resolution network for multiple
  degradations.
\newblock \emph{Proceedings of the IEEE/CVF Conference on Computer Vision and
  Pattern Recognition (CVPR)}, pages 3262--3271, 2018.

\bibitem[Zhang et~al.(2024)Zhang, Zeng, Pan, Shen, and Chen]{LLEMamba}
Xuanqi Zhang, Haijin Zeng, Jinwang Pan, Qiangqiang Shen, and Yongyong Chen.
\newblock Llemamba: Low-light enhancement via relighting-guided mamba with deep
  unfolding network.
\newblock \emph{arXiv preprint arXiv:2406.01028}, pages 1--12, 2024.

\bibitem[Zhang et~al.(2019)Zhang, Zhang, and Guo]{Kind}
Yonghua Zhang, Jiawan Zhang, and Xiaojie Guo.
\newblock Kindling the darkness: A practical low-light image enhancer.
\newblock \emph{Proceedings of the ACM International Conference on MultiMedia},
  pages 1632--1640, 2019.

\bibitem[Zhang et~al.(2021)Zhang, Guo, Ma, Liu, and Zhang]{Kind_plus}
Yonghua Zhang, Xiaojie Guo, Jiayi Ma, Wei Liu, and Jiawan Zhang.
\newblock Beyond brightening low-light images.
\newblock \emph{International Journal of Computer Vision}, pages 1013--1037,
  2021.

\bibitem[Zhou et~al.(2020)Zhou, Hou, Chen, Feng, and Yan]{efficient-bottleneck}
Daquan Zhou, Qibin Hou, Yunpeng Chen, Jiashi Feng, and Shuicheng Yan.
\newblock Rethinking bottleneck structure for efficient mobile network design.
\newblock \emph{European Conference on Computer Vision (ECCV)}, pages 680--697,
  2020.

\bibitem[Zhou et~al.(2024)Zhou, Dong, Liu, Liu, Min, Zhai, and Chen]{GLARE}
Han Zhou, Wei Dong, Xiaohong Liu, Shuaicheng Liu, Xiongkuo Min, Guangtao Zhai,
  and Jun Chen.
\newblock Glare: Low light image enhancement via generative latent feature
  based codebook retrieval.
\newblock In \emph{Proceedings of the European Conference on Computer Vision
  (ECCV)}, pages 1--19, 2024.

\end{thebibliography}
